\documentclass[10pt,twocolumn,letterpaper]{article}

\usepackage[pagenumbers]{cvpr} 

\definecolor{cvprblue}{rgb}{0.21,0.49,0.74}
\usepackage[accsupp]{axessibility} 
\usepackage[pagebackref,breaklinks,colorlinks,allcolors=cvprblue]{hyperref}
\usepackage{color, colortbl}
\usepackage{multirow}
\usepackage{comment}
\definecolor{Gray}{rgb}{0.9,0.9,0.9}

\definecolor{sotaa}{rgb}{1,0,0}
\newcommand{\sotaa}[1]{\textcolor{sotaa}{#1}}

\definecolor{sotab}{rgb}{0,0,1}
\newcommand{\sotab}[1]{\textcolor{sotab}{#1}}

\def\paperID{870} 
\def\confName{CVPR}
\def\confYear{2025}

\title{Progressive Focused Transformer for Single Image Super-Resolution}

\author{
Wei Long\textsuperscript{1}, Xingyu Zhou\textsuperscript{1}, Leheng Zhang\textsuperscript{1}, Shuhang Gu\textsuperscript{1*} \\
\textsuperscript{1}University of Electronic Science and Technology of China \\
{\tt\small weilong@std.uestc.edu.cn, shuhanggu@gmail.com} \\
{\tt\small \url{https://github.com/LabShuHangGU/PFT-SR}}
}

\begin{document}
\maketitle

\renewcommand{\thefootnote}{\fnsymbol{footnote}}
\footnotetext[1]{corresponding author}

\begin{abstract}
Transformer-based methods have achieved remarkable results in image super-resolution tasks because they can capture non-local dependencies in low-quality input images. However, this feature-intensive modeling approach is computationally expensive because it calculates the similarities between numerous features that are irrelevant to the query features when obtaining attention weights. These unnecessary similarity calculations not only degrade the reconstruction performance but also introduce significant computational overhead. How to accurately identify the features that are important to the current query features and avoid similarity calculations between irrelevant features remains an urgent problem. To address this issue, we propose a novel and effective \textbf{Progressive Focused Transformer (PFT)} that links all isolated attention maps in the network through \textbf{Progressive Focused Attention (PFA)} to focus attention on the most important tokens. PFA not only enables the network to capture more critical similar features, but also significantly reduces the computational cost of the overall network by filtering out irrelevant features before calculating similarities. Extensive experiments demonstrate the effectiveness of the proposed method, achieving state-of-the-art performance on various single image super-resolution benchmarks.

\end{abstract}    
\section{Introduction}
\label{sec:intro}

\begin{figure}
    \centering
    \includegraphics[width=\linewidth]{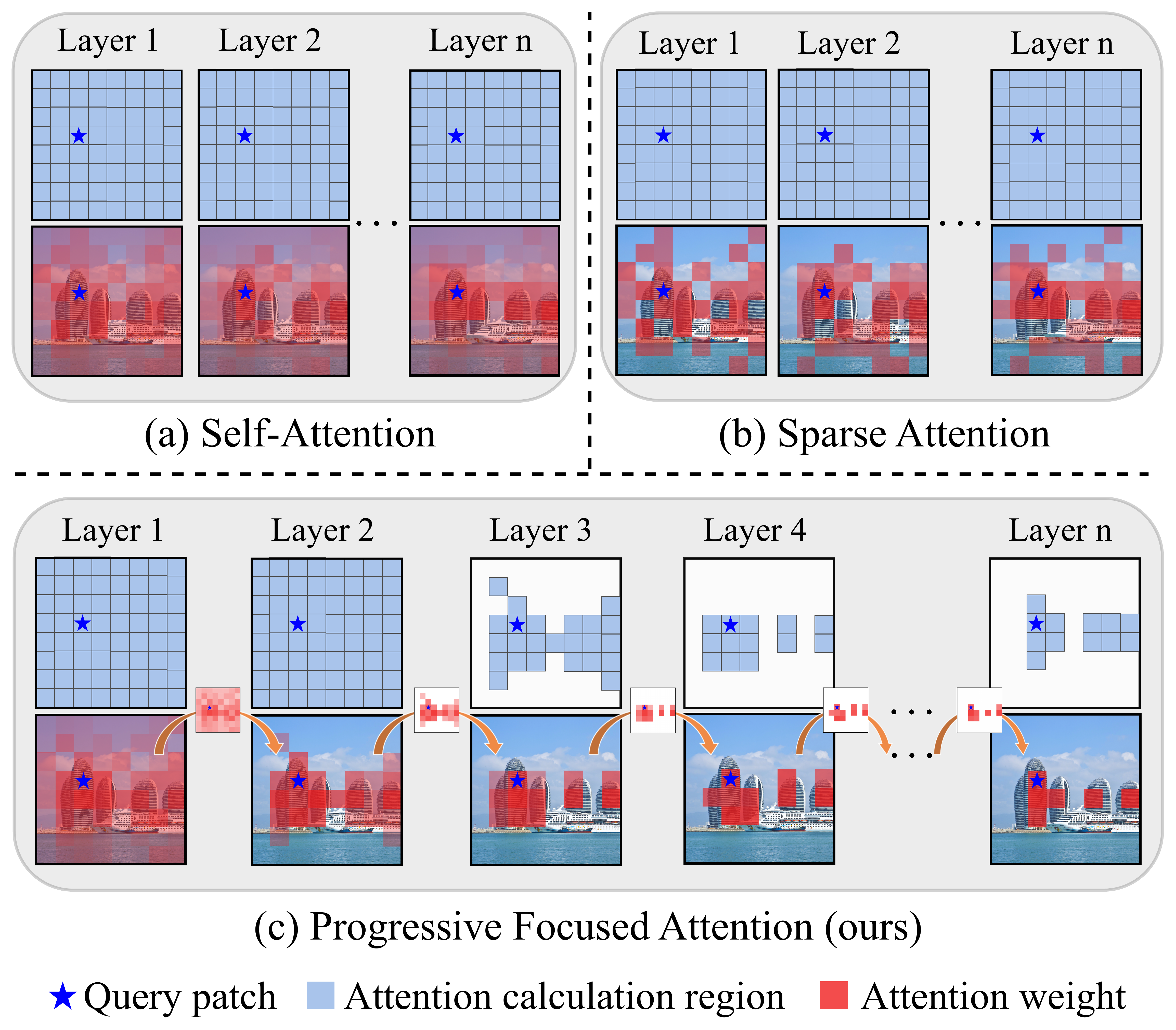}
        \vspace{-3mm}
    \caption{Comparison of different attention mechanisms. (a) Vanilla self-attention calculates attention weights in the whole window and generates non-zero weights for both highly relevant and less relevant tokens.  (b) Sparse Attention is able to filter out the impact of less relevant tokens with small weights, but still requires the calculation of attention weights for all tokens.  (c) Progressive Focused Attention connects isolated attention maps, leveraging attention weights to skip unnecessary computations and better aggregate relevant tokens. }
    \label{fig:motivation}
    \vspace{-3mm}
\end{figure}

Image Super-Resolution (SR) aims to reconstruct high-resolution details and textures from Low-Resolution (LR) images. 
Effective SR techniques not only enhance visual quality but also enable more meaningful information extraction from high-resolution outputs.
It has critical applications in fields such as medical imaging, satellite remote sensing, and video surveillance.

Several classical Convolutional Neural Networks (CNNs)~\cite{dong2014learning, dong2015image, gu2015convolutional, lim2017enhanced} designed for SR have achieved promising reconstruction results.~Due to the limitations of convolution operations, these methods can capture only local feature information.~Many studies have focused on enabling networks to achieve broader feature interaction to improve image restoration.~Recently, Transformer-based methods~\cite{liang2021swinir, chen2023activating, zhang2022efficient, zhou2023srformer} have been increasingly applied to image reconstruction, owing to their ability to effectively leverage long-range dependencies, thereby enhancing the recovery of detailed information in images.

Transformer-based super-resolution methods utilize cascaded self-attention blocks to recover the missing High-Resolution (HR) details.
At the core of the self-attention block is the attention map, which determines how the input features are aggregated to generate output features.
Due to the quadratic computational complexity of the self-attention mechanism, most of the existing methods confine attention computation within local windows and only aggregate a limited number of tokens to update image features.
Therefore, how to exploit information of relevant tokens while alleviating the disruption from irrelevant tokens is critical in Transformer-based SR models.
For the pursuit of high-quality SR results, one category of research tries to enrich the information source by enlarging the window size~\cite{chen2023activating, zhou2023srformer}, leveraging cross-window information~\cite{NEURIPS2022_a37fea8e}, and introducing an external token dictionary~\cite{Zhang_2024_CVPR}.
Since more information can be incorporated into the weighted feature updating process, these methods have shown promising capability in delivering superior SR results.
However, calculating similarities between more tokens inevitably leads to extra computational footprint. These methods still need to compromise on model overhead and introduce limited extra information.
Another category of studies investigates sophisticated weighting strategies to make better use of input information~\cite{Mei_2021_CVPR,chen2023learning}.
Instead of combining all the tokens according to the normalized productive similarities, as in a vanilla self-attention block, various elaborately designed weighting mechanisms have been exploited to make better use of highly relevant tokens while at the same time filtering out less relevant tokens.
%
While the overall concept of seeking more rational weights is highly appealing, the challenge of obtaining enhanced attention weights remains an open problem in the literature.

\vspace{-0.4em}
In this paper, we link the attention maps between adjacent layers and propose a Progressive Focused Attention (PFA) block to better leverage input tokens across a wider range.
Instead of combining input features merely according to the calculated attention map $\mathbf{A}^l_{cal}$, we inherit attention information from the previous block in a productive manner $\mathbf{A}^l = Norm\left(\mathbf{A}^{l-1} \odot \mathbf{A}^l_{cal}\right)$, where $Norm(\cdot)$ is the row-wise normalization operation.
Such an attention inheritance mechanism enables our method to make a comprehensive measurement of token similarity: tokens that are consistently similar to each other are empowered with larger weights to help each other while the weights of irrelevant tokens are suppressed during the cascaded multiplications.
Furthermore, in addition to boosting weights for important tokens and suppressing weights for less relevant tokens, another advantage of our PFA lies in its calculation pre-filtering capability.
Specifically, since the final attention map $\mathbf{A}^l $ will be multiplied by the previous attention map $ \mathbf{A}^{l-1}$, small weights in $ \mathbf{A}^{l-1}$ are a sign of unimportant tokens, for which the weight calculation process can be skipped.
As shown in Fig.\ref{fig:motivation}, the previous attention map not only helps our PFA block to generate better weights which emphasize highly relevant tokens, but also enables it to skip unnecessary similarity calculations so that we are able to afford the computational overhead of large window.
%
%
Therefore, with comparable computational resources, our SR model could employ a larger window size to leverage input tokens from a broader range.
%


The main contributions of this paper are as follows:

\begin{itemize}
\item 
We propose the idea of Progressive Focused Attention (PFA) which inherits attention weights from previous layers in a productive manner. PFA not only avoids unnecessary similarity computation with less relevant tokens but also endows highly relevant tokens with larger weights to better enhance the input feature. 
\item
We instantiate a Progressive Focused Transformer (PFT) for image super-resolution, which takes the progressive focusing benefits of PFA and systematically arranges computational resources for different layers.
%
%
~This property enables the PFT to leverage the merit of large window while maintaining model efficiency.

\item
We conduct extensive experiments on super-resolution benchmarks. Our superior experimental results over recent state-of-the-art methods as well as our detailed ablation studies demonstrate the effectiveness of our method.
\end{itemize}

\section{Related Work}
\label{sec:related}
\noindent \textbf{Transformer-based SR.} Transformer models were initially used for natural language processing~\cite{vaswani2017attention, bert}. 
Subsequently, they have attracted extensive research interest in the field of computer vision~\cite{dosovitskiy2020image, cao2021video, yang2020learning, lu2022transformer, fang2022hybrid}.
Due to its excellent feature extraction capabilities, Transformers have also been introduced to low-level visual tasks, including single image super-resolution~\cite{li2023efficient, zhang2024hit, yoo2023enriched}. 
Classical SR models~\cite{tai2017image, kim2016deeply, kim2016accurate, zhang2018learning} based on CNN architectures were limited by narrow receptive fields and limited feature interaction capabilities, and their effectiveness was gradually surpassed by Transformer-based approaches. 
For example, IPT~\cite{Chen_2021_CVPR} implements a pre-trained Transformer-based model for better super-resolution. 
SwinIR~\cite{liang2021swinir} improves image quality by using shifted window attention and residual blocks, effectively capturing a broad range of features for restoration.
Subsequently, CAT~\cite{NEURIPS2022_a37fea8e} improves image restoration by using rectangle-window self-attention and axial shifting to enhance cross-window interaction.
To address the perceptual limitations of window-based methods, several approaches have been proposed.
HAT~\cite{chen2023activating} combines channel and window-based attention to overcome these limitations and enhance cross-window interactions for superior image reconstruction. 
Another approach is ATD~\cite{Zhang_2024_CVPR}, which addresses the limitations of local windows by using an Adaptive-Token Dictionary and category-based self-attention to incorporate global information and enhance features.~In addition, IPG~\cite{Tian_2024_CVPR} improves SR by prioritizing detail-rich pixels and using flexible local and global graphs to enhance reconstruction.
%
Unlike existing methods, we propose a PFT method that can better focus on key tokens and filter out irrelevant features through PFA, thereby reducing the impact of irrelevant features on SR results and the computational effort.

\vspace{0.5mm}
\noindent \textbf{Transformer with Sparse Attention.} The dense attention maps in Transformers may introduce features from irrelevant locations, which interferes with the effectiveness of feature extraction. 
To address this, many approaches explore how to leverage sparse attention calculations to better utilize relevant information. 
For example, NLSA~\cite{Mei_2021_CVPR} combines non-local operations with sparse representations, using dynamic sparse attention to focus on pertinent features, thereby improving efficiency and robustness while reducing computational costs. 
ART~\cite{zhang2023accurate} enhances image restoration by combining both dense and sparse attention mechanisms, expanding the receptive field to enable more effective interactions and better feature representation. 
Additionally, a class of methods~\cite{zhao2019explicit, wang2022kvt} explicitly selects the most relevant positions in the attention map for retention, filtering out irrelevant information that might interfere with feature aggregation.
Building on this, DRSformer~\cite{chen2023learning} further develops the idea of sparse selection by introducing a learnable top-$k$ selection operator, that adaptively retains the most relevant self-attention values, thus improving image restoration. 
We propose the PFT method, which connects isolated attention maps across the entire network to efficiently identify the positions each attention map collectively focuses on. 
PFA performs a Hadamard product on adjacent attention maps to pinpoint the most relevant attention locations, thereby enhancing the filtering of irrelevant information.

\section{Progressive Focused Transformer}
\label{sec:method}

\subsection{Preliminaries}
\label{Previous Unfocused Attention}

\noindent \textbf{Self-Attention.} 
The self-attention mechanism~\cite{vaswani2017attention} is a fundamental part of Transformer, which computes the similarity between tokens and aggregates features based on the corresponding attention weights.

Given the query matrix $\mathbf{Q} \in \mathbb{R}^{N\times d}$, key matrix $\mathbf{K} \in \mathbb{R}^{N\times d}$,
the attention weights are calculated as follows:
\begin{align}
\label{eq:sa}
	\mathbf{A}_{sa} &= \text{Softmax} \left( \mathbf{Q} \mathbf{K}^\top/{\sqrt{d}} \right),
\end{align}
where the $i$-th row in $\mathbf{A}_{sa}$ represents normalized weights to aggregate value tokens $\mathbf{V} \in \mathbb{R}^{N\times d}$ and the output of self-attention module is obtained by $\mathbf{O}_{SA}=\mathbf{A}_{sa}\mathbf{K}$.
Although the exponential function in Softmax could suppress irrelevant tokens to some extent, the above vanilla self-attention still apply dense weights which utilize all the input tokens in the window to generate the final output.
Moreover, another feature of the above self-attention lies in its computational overhead.
The complexity of Eq.~\eqref{eq:sa} grows quadratically with the number of tokens $N$. Therefore, most existing methods need to perform self-attention calculations in small windows to manage this computational burden.

\vspace{2mm}
\noindent \textbf{Sparse Attention.} To mitigate the negative impact of irrelevant features in the dense attention map, numerous sparse attention methods have been proposed. 
Generally speaking, the overall idea of sparse attention is to introduce a selection operation to sparsify the attention weights:
\begin{align}
\label{eq:ssa}
	\mathbf{A}_{ssa} &= \text{Softmax}\left(\mathcal{S}\left(\mathbf{Q} \mathbf{K}^\top/\sqrt{d} \right)\right),
\end{align}
where $\mathcal{S}(\cdot)$ is a sparsification operation which removes the weights for less relevant tokens.~The most commonly adopted sparsify operation is the top-$k$ selection, which only allows $K$ non-zero values in each row.
Despite its strong capability of irrelevant token suppression, top-$k$ selection only adjusts aggregation weights according to a single step of similarity calculation and has limited ability to make effective use of highly relevant tokens.

\begin{figure*}
    \centering
    \includegraphics[width=\linewidth]{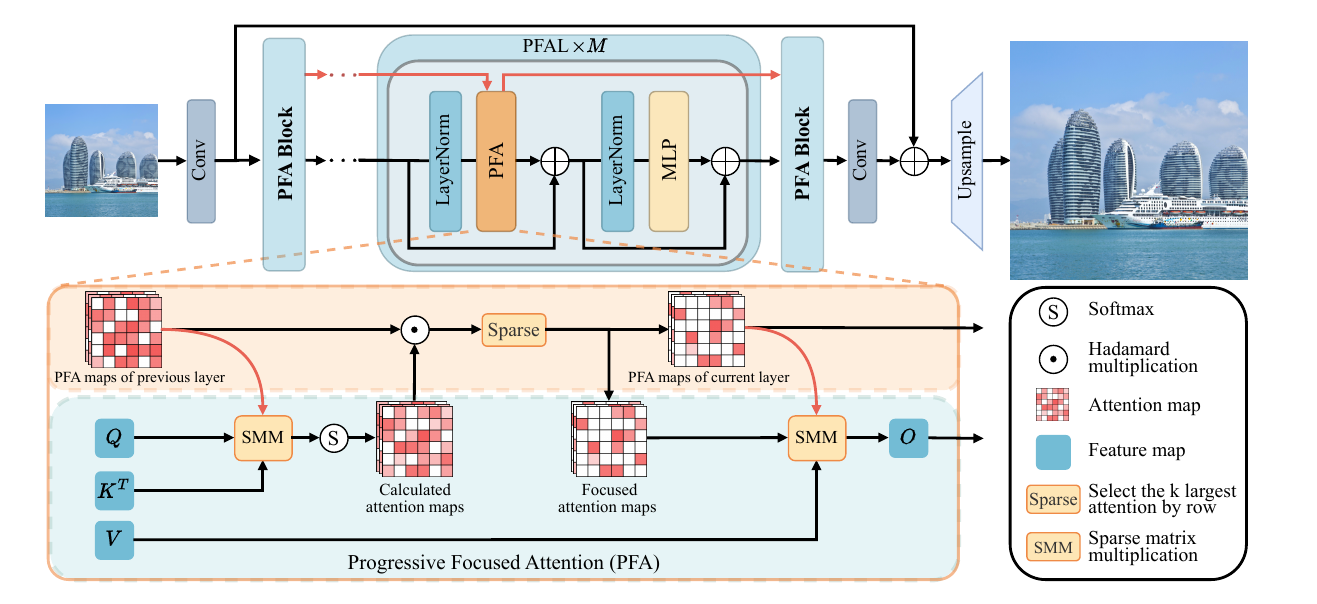}
    \caption{The overall architecture of PFT. PFA Block consists of $M$ Progressive Focused Attention Layers (PFAL). Each PFA takes both image features and aggregated PFA maps from previous layers as input. Sparse Matrix Multiplication (SMM) ensures each row of $Q$ interacts only with sparse columns of $K^T$, producing calculated attention maps. After applying the Hadamard product and sparse focusing, PFA maps of the current layer are obtained and used with the $V$ matrix in an SMM operation to generate attention-aggregated features.}
    \label{fig:PFT}
    \vspace{-3mm}
\end{figure*}

\vspace{2mm}
\subsection{Progressive Focused Attention}
\label{PFA}
As reviewed in the previous section, although sparse attention methods are able to improve the vanilla self-attention by alleviating the negative impact of irrelevant tokens, they still have limited capability in further distinguishing highly relevant tokens from mildly relevant tokens.
Moreover, both the vanilla and sparse attention methods fail to identify the relevant feature regions prior to similarity calculation, resulting in a significant amount of unnecessary calculations, which impedes the incorporation of additional input tokens.

In this paper, we propose Progressive Focused Attention, which links the attention maps between adjacent layers to better aggregate relevant features.
Generally, the proposed PFA advances the existing methods in two aspects.
Firstly, PFA makes use of attention maps from previous layers to better weight input features. The weights of highly relevant tokens that are consistently similar to each other across layers will be enhanced during the cascade multiplication process.
Secondly, PFA leverages the attention map of previous layers to identify irrelevant features for saving computational overhead. This selective calculation mechanism enables us to conduct self-attention within a larger window.
%
In the remaining part of this section, we first introduce the basic idea of our PFA, which inherits the attention map from the previous layers through a Hadamard product.
Then, we introduce how we could further leverage previous attention maps to reduce computational overhead.

\vspace{1mm}
\noindent \textbf{Progressive Attention Across Layers.} 
As introduced in Eq.~\eqref{eq:sa} and~\eqref{eq:ssa}, the previous methods obtain attention weights merely according to the calculated similarities between tokens from the current layer.
The standard similarity calculation has limited discriminative capability in boosting highly relevant tokens while suppressing less relevant tokens.
To address this issue, we propose to inherit attention weights from previous layers in a productive manner:
\begin{align}
\label{eq:crosslayer_attention}
    \mathbf{A}^{l} &= Norm(\mathbf{A}^{l-1} \odot \mathbf{A}^{l}_{cal}),
\end{align}
where $\mathbf{A}^{l}_{cal}$ is the calculated attention map for the $l$-th layer according to the vanilla self-attention mechanism as in Eq.~\eqref{eq:sa}; 
$\mathbf{A}^{l-1}$ is the attention map of the previous layer; $Norm(\cdot)$ represents the row-wise normalization operation, which ensures the summation of each row to be 1.
As indicated in Eq.~\eqref{eq:crosslayer_attention}, the final attention map in the $l$-th layer is obtained by multiplying the calculated similarities with the attention map from the previous layer.
Such a mechanism enables us to make a comprehensive assessment of token similarity: the weights for less relevant tokens will be wiped out during the successive multiplication process, while important tokens with high similarity values across layers will be assigned with larger weights after normalization.
As will be validated in our ablation studies, with improved feature aggregation weights, 
such a Progressive Attention mechanism could deliver better SR results over the top-$k$ sparse attention as well as the standard self-attention block.

\noindent
\textbf{Progressive Focused Attention.}
Since all the normalized similarities are less than 1, the above strategy of multiplicative attention inheritance could 
gradually filter out the impact of irrelevant tokens and obtain sparse weights as the network depth increases.
Actually, based on the previous attention map $\mathbf{A}^{l-1}$, we are able to identify unimportant positions before calculation.
Therefore, we can avoid unnecessary computations with a Sparse Matrix Multiplication (SMM) operation:
\begin{equation} 
    \label{eq:sparse_sa_calculation}
    \mathbf{A}^{l}_{sc} = Softmax(\mathbf{\Psi} (\mathbf{Q}^{l}, (\mathbf{K}^{l})^\top, \mathbf{I}^{l-1})),
\end{equation}
where $\mathbf{\Psi}$ denotes the SMM operation; $\mathbf{I}^{l-1} \in \mathbb{R}^{N\times N}$ is a sparse index matrix;  we calculate $\mathbf{A}^{l}_{sc}(i,j)$  with $\mathbf{Q}^{l}(i,:)$ and ${\mathbf{K}^{l}(j,:)}^T$ only if $\mathbf{I}^{l-1}(i,j) = 1$, otherwise we set $\mathbf{A}^{l}_{sc}(i,j)=0$.
Note that after the SMM operation we still need to normalize each row of the calculation results using the Softmax function.
For an all-ones indexing matrix, the SMM operation in Eq.~\eqref{eq:sparse_sa_calculation} is equivalent to the standard attention calculation process as in Eq.~\eqref{eq:sa}; However, if we are able to obtain a sparse $\mathbf{I}^{l-1}$ prior to the calculation process, a significant portion of computations can be skipped.
Due to our progressive attention mechanism, we utilize $\mathbf{A}^{l-1}$ to indicate skippable positions.
Specifically, inspired by the success of top-$k$ sparse attention, we only keep $K^l$ non-zero attention weights for each layer and generate the indexing matrix $\mathbf{I}^{l}$ according to the non-zero positions in $\mathbf{A}^{l}$.
To sum up, the attention map of our progressive focused attention is achieved through the following formula:
\begin{align}
\label{eq:pfa}
    \begin{cases} 
    \mathbf{A}^{l}_{sc}\!\!\!\!\!&= ~Softmax(\mathbf{\Psi} (\mathbf{Q}^{l}, (\mathbf{K}^{l})^\top, \mathbf{I}^{l-1})), \\
    \mathbf{A}^{l}\!&= ~\mathbf{\mathcal{S}}_{K^l}(Norm(\mathbf{A}^{l}_{sc} \odot \mathbf{A}^{l-1})),   \\
    \mathbf{I}^{l}\!&=  ~\operatorname{Sign}(\mathbf{A}^{l}),
    \end{cases}
\end{align}
%
Where  $\mathbf{\mathcal{S}}_{K^l}(\cdot)$ is the sparsification operation that selects the top $K^l$ values in each row and $\operatorname{Sign}(\cdot)$ generates a binary index matrix according to the sign of input matrix.
Then, we can also use the SMM operation to calculate the output matrix of the $l$-th layer:
\begin{equation} 
    \label{eq:attn_v}
    \mathbf{O}^{l} = \mathbf{\Psi} (\mathbf{A}^{l}, \mathbf{V}^{l}, \mathbf{I}^{l}).
\end{equation}

For the first attention block, in which we do not have previous attention map, we simply set $\mathbf{A}^{0}$ and $\mathbf{I}^{0}$ as all-ones matrix and calculate $\mathbf{A}^{1}$ as standard self-attention in Eq.~\eqref{eq:sa} to aggregate input features.
For subsequent layers, we  exploit the attention map from previous layer to generate better weights and filter out unnecessary computations as in Eq.~\eqref{eq:pfa}.
Since tokens which have been determined as irrelevant tokens to a specific query token in previous layers will not be considered in the following layers at all, to avoid filtering-out important tokens at early stages, we set $K^1 = N$ and gradually reduce the nonzero values with a focus ratio: $K^l = \alpha K^{l-1}$, where $K$ represents the number of retained attention values within the window.
With $0<\alpha<1$, our cascade PFA layers gradually focus on important candidates and reduce the number of key-tokens to be considered for each input query-token, resulting in significant reduction in computational burden for each window. Notably, using standard matrix multiplication to implement the SMM operation is highly inefficient. Therefore, we developed efficient CUDA kernels for SMM. This implementation significantly enhances the efficiency of our model, enabling PFT to achieve inference speeds comparable to those of existing methods.

\subsection{The Overall Network Architecture}

As illustrated in Fig.~\ref{fig:PFT}, the overall architecture of PFT follows the structure adopted in recent state-of-the-art Transformer-based SR models~\cite{liang2021swinir, chen2023activating, Zhang_2024_CVPR}. 
While, we replace the standard self-attention blocks with our proposed PFA block for the pursuit of better SR results.
Notably, we also incorporate the shifted window approach from the Swin Transformer, though it is omitted in the illustration for simplicity.
Thus, the attention map in PFA is passed from the first layer to the last through two pathways, divided by the shift operation into even and odd layers for focused processing. 
Additionally, we incorporate LePE positional encoding~\cite{dong2022cswin} during attention computation.
For classical SR, the PFT network consists of 6 blocks, with the number of attention layers in each block being [4, 4, 4, 6, 6, 6]. The model utilizes multi-head attention with 6 attention heads, and the total number of channels is 240. The window size is 32$\times$32, and the number of retained attention values in each block is [1024, 256, 128, 64, 32, 16], respectively. For lightweight SR, the PFT-light network is composed of five blocks, with the number of attention layers in each block being [2, 4, 6, 6, 6]. It employs 4 attention heads, with a total of 52 channels. The window size is 32$\times$32, and the number of retained attention values in each block follows the pattern [1024, 256, 128, 64, 32].

\vspace{1mm}
\noindent \textbf{Analysis of Computational Complexity.} Self-Attention and sparse attention have the same computational cost, as both measure similarity with neighboring positions by interacting with tokens within non-overlapping windows. When the input feature map size is $h \times w \times C$ and the window size is $W \times W$, the feature map is divided into $\frac{h}{W} \times \frac{w}{W}$ windows, each with $W \times W$ tokens. Assuming that the network has a total of $L$ attention layers, the computational complexity of the attention component for the entire network is: 

\vspace{-2mm}
\begin{equation}
	\Omega(SA) = 4 h w L C^2 + 2 W^2 h w L C.
\end{equation}
PFA can control the proportion of key tokens that are most relevant to the query features by adjusting the retention ratio $\alpha$. The index matrix $\mathbf{I}$ allows shallow attention aggregation information to be carried into deeper layers for further focusing, resulting in a gradually decreasing $K$.
For a network with \( L \) attention layers, we assume that the number of retained attention values within the window decreases, as expressed by the following formula: \( K^l = \alpha K^{l-1} \), where \( 0 < \alpha < 1 \) and \( K^1 \) is generally set to \( W^2 \). It can be represented as \( K^l = W^2 \alpha^{l-1} \). The overall computational complexity of the model is:

\vspace{-2mm}
\begin{equation}
\Omega(PFA) = \sum_{l=1}^{L} \left( 4 h w C^2 + 2 \alpha^{(l-1)} W^2 h w C \right),
\end{equation}
\vspace{-2mm}

\noindent where $\alpha^{(l-1)}$ controls the decay of the focus ratio across layers, with $0 < \alpha < 1$. As $l$ increases, the focus ratio progressively decreases. The first term, $4 h w C^2$, represents the computational complexity of the projection layers. The second term, $2 \alpha^{(l-1)} W^2 h w C$, accounts for the complexity of computing the similarity between the query and the selected key tokens, as well as redistributing the corresponding value information at the focused positions. Obviously, since $\alpha$ is less than 1, the complexity of this term decreases exponentially with depth. For example, with $\alpha = 0.5$, the computational complexity is reduced to just 6.25\% of the original after four decay steps.

\section{Experiments}
\label{sec:experiments}

\begin{table*}[htbp]
\captionsetup{font={small}}
\scriptsize
\vspace{-2mm}
  \begin{center}
  \begin{tabular}{|p{1.8cm}|c|c|c|cc|cc|cc|cc|cc|}
    \hline
    \multirow{2}{*}{\textbf{Method}} & \multirow{2}{*}{\textbf{Scale}} & \multirow{2}{*}{\textbf{Params}} & \multirow{2}{*}{\textbf{FLOPs}} & \multicolumn{2}{c|}{\textbf{Set5}} & \multicolumn{2}{c|}{\textbf{Set14}} & \multicolumn{2}{c|}{\textbf{BSD100}} & \multicolumn{2}{c|}{\textbf{Urban100}} & \multicolumn{2}{c|}{\textbf{Manga109}} \\

    & & & & PSNR & SSIM & PSNR & SSIM & PSNR & SSIM & PSNR & SSIM & PSNR & SSIM   \\

    \hline
    EDSR~\cite{lim2017enhanced}       & $\times$2 & 42.6M & 22.14T  & 38.11 & 0.9602 & 33.92 & 0.9195 & 32.32 & 0.9013 & 32.93 & 0.9351 & 39.10 & 0.9773 \\
    RCAN~\cite{zhang2018image}        & $\times$2 & 15.4M & 7.02T   & 38.27 & 0.9614 & 34.12 & 0.9216 & 32.41 & 0.9027 & 33.34 & 0.9384 & 39.44 & 0.9786 \\
    HAN~\cite{niu2020single}          & $\times$2 & 63.6M & 7.24T   & 38.27 & 0.9614 & 34.16 & 0.9217 & 32.41 & 0.9027 & 33.35 & 0.9385 & 39.46 & 0.9785 \\
    IPT~\cite{Chen_2021_CVPR}         & $\times$2 & 115M  & 7.38T   & 38.37 & - & 34.43 & - & 32.48 & - & 33.76 & - & - & - \\
    SwinIR~\cite{liang2021swinir}     & $\times$2 & 11.8M & 3.04T   & 38.42 & 0.9623 & 34.46 & 0.9250 & 32.53 & 0.9041 & 33.81 & 0.9433 & 39.92 & 0.9797 \\
    CAT-A~\cite{NEURIPS2022_a37fea8e} & $\times$2 & 16.5M & 5.08T   & 38.51 & 0.9626 & 34.78 & 0.9265 & 32.59 & 0.9047 & 34.26 & 0.9440 & 40.10 & 0.9805 \\
    ART~\cite{zhang2023accurate}      & $\times$2 & 16.4M & 7.04T   & 38.56 & 0.9629 & 34.59 & 0.9267 & 32.58 & 0.9048 & 34.30 & 0.9452 & 40.24 & 0.9808 \\
    HAT~\cite{chen2023activating}     & $\times$2 & 20.6M & 5.81T   & \sotab{38.63} & 0.9630 & 34.86 & 0.9274 & 32.62 & 0.9053 & 34.45 & 0.9466 & 40.26 & 0.9809 \\
    IPG~\cite{Tian_2024_CVPR}         & $\times$2 & 18.1M & 5.35T   & 38.61 & \sotab{0.9632} & 34.73 & 0.9270 & 32.60 & 0.9052 & 34.48 & 0.9464 & 40.24 & \sotab{0.9810} \\
    ATD~\cite{Zhang_2024_CVPR}        & $\times$2 & 20.1M & 6.07T   & 38.61 & 0.9629 & \sotab{34.95} & \sotab{0.9276} & \sotab{32.65} & \sotab{0.9056} & \sotab{34.70} & \sotab{0.9476} & \sotab{40.37} & \sotab{0.9810} \\

    \rowcolor{Gray}
    \textbf{PFT} (Ours)               & $\times$2 & 19.6M & 5.03T   & \sotaa{38.68} & \sotaa{0.9635} &  \sotaa{35.00} &  \sotaa{0.9280} & \sotaa{32.67} & \sotaa{0.9058} & \sotaa{34.90} & \sotaa{0.9490} & \sotaa{40.49} & \sotaa{0.9815} \\

    \hline \hline
    EDSR~\cite{lim2017enhanced}       & $\times$3 & 43.0M & 9.82T & 34.65 & 0.9280 & 30.52 & 0.8462 & 29.25 & 0.8093 & 28.80 & 0.8653 & 34.17 & 0.9476 \\
    RCAN~\cite{zhang2018image}        & $\times$3 & 15.6M & 3.12T & 34.74 & 0.9299 & 30.65 & 0.8482 & 29.32 & 0.8111 & 29.09 & 0.8702 & 34.44 & 0.9499 \\
    HAN~\cite{niu2020single}          & $\times$3 & 64.2M & 3.21T & 34.75 & 0.9299 & 30.67 & 0.8483 & 29.32 & 0.8110 & 29.10 & 0.8705 & 34.48 & 0.9500 \\
    IPT~\cite{Chen_2021_CVPR}         & $\times$3 & 116M  & 3.28T & 34.81 & -      & 30.85 & -      & 29.38 & -      & 29.49 & -      & -     & -      \\
    SwinIR~\cite{liang2021swinir}     & $\times$3 & 11.9M & 1.35T & 34.97 & 0.9318 & 30.93 & 0.8534 & 29.46 & 0.8145 & 29.75 & 0.8826 & 35.12 & 0.9537 \\
    CAT-A~\cite{NEURIPS2022_a37fea8e} & $\times$3 & 16.6M & 2.26T & 35.06 & 0.9326 & 31.04 & 0.8538 & 29.52 & 0.8160 & 30.12 & 0.8862 & 35.38 & 0.9546 \\
    ART~\cite{zhang2023accurate}      & $\times$3 & 16.6M & 3.12T & 35.07 & 0.9325 & 31.02 & 0.8541 & 29.51 & 0.8159 & 30.10 & 0.8871 & 35.39 & 0.9548 \\
    HAT~\cite{chen2023activating}     & $\times$3 & 20.8M & 2.58T & 35.07 & 0.9329 & 31.08 & 0.8555 & 29.54 & 0.8167 & 30.23 & 0.8896 & 35.53 & 0.9552 \\
    IPG~\cite{Tian_2024_CVPR}         & $\times$3 & 18.3M & 2.39T & 35.10 & \sotab{0.9332} & 31.10 & 0.8554 & 29.53 & 0.8168 & 30.36 & 0.8901 & 35.53 & 0.9554 \\
    ATD~\cite{Zhang_2024_CVPR}        & $\times$3 & 20.3M & 2.69T & \sotab{35.11} & 0.9330 & \sotab{31.13} & \sotab{0.8556} & \sotab{29.57} & \sotab{0.8176} & \sotab{30.46} & \sotab{0.8917} & \sotab{35.63} & \sotab{0.9558} \\

    \rowcolor{Gray}
    \textbf{PFT} (Ours)              & $\times$3 & 19.8M & 2.23T & \sotaa{35.15}  & \sotaa{0.9333}   & \sotaa{31.16}   & \sotaa{0.8561} & \sotaa{29.58} & \sotaa{0.8178} & \sotaa{30.56} & \sotaa{0.8931} & \sotaa{35.67} & \sotaa{0.9560} \\

    \hline \hline
    EDSR~\cite{lim2017enhanced}       & $\times$4 & 43.0M & 5.54T & 32.46 & 0.8968 & 28.80 & 0.7876 & 27.71 & 0.7420 & 26.64 & 0.8033 & 31.02 & 0.9148 \\
    RCAN~\cite{zhang2018image}        & $\times$4 & 15.6M & 1.76T & 32.63 & 0.9002 & 28.87 & 0.7889 & 27.77 & 0.7436 & 26.82 & 0.8087 & 31.22 & 0.9173 \\
    HAN~\cite{niu2020single}          & $\times$4 & 64.2M & 1.81T & 32.64 & 0.9002 & 28.90 & 0.7890 & 27.80 & 0.7442 & 26.85 & 0.8094 & 31.42 & 0.9177 \\
    IPT~\cite{Chen_2021_CVPR}         & $\times$4 & 116M  & 1.85T & 32.64 & -      & 29.01 & -      & 27.82 & -      & 27.26 & -      & -     & -      \\   
    SwinIR~\cite{liang2021swinir}     & $\times$4 & 11.9M & 0.76T & 32.92 & 0.9044 & 29.09 & 0.7950 & 27.92 & 0.7489 & 27.45 & 0.8254 & 32.03 & 0.9260 \\
    CAT-A~\cite{NEURIPS2022_a37fea8e} & $\times$4 & 16.6M & 1.27T & 33.08 & 0.9052 & 29.18 & 0.7960 & 27.99 & 0.7510 & 27.89 & 0.8339 & 32.39 & 0.9285 \\
    ART~\cite{zhang2023accurate}      & $\times$4 & 16.6M & 1.76T & 33.04 & 0.9051 & 29.16 & 0.7958 & 27.97 & 0.7510 & 27.77 & 0.8321 & 32.31 & 0.9283 \\
    HAT~\cite{chen2023activating}     & $\times$4 & 20.8M & 1.45T & 33.04 & 0.9056 & 29.23 & 0.7973 & 28.00 & 0.7517 & 27.97 & 0.8368 & 32.48 & 0.9292 \\
    IPG~\cite{Tian_2024_CVPR}         & $\times$4 & 17.0M & 1.30T & \sotaa{33.15} & \sotab{0.9062} & \sotab{29.24} & 0.7973 & 27.99 & 0.7519 & 28.13 & 0.8392 & 32.53 & \sotab{0.9300} \\
    ATD~\cite{Zhang_2024_CVPR}        & $\times$4 & 20.3M & 1.52T & \sotab{33.10} & 0.9058 & \sotab{29.24} & \sotab{0.7974} & \sotab{28.01} & \sotab{0.7526} & \sotab{28.17} & \sotab{0.8404} & \sotab{32.62} & \sotaa{0.9306} \\
    
    \rowcolor{Gray}
    \textbf{PFT} (Ours)               & $\times$4 & 19.8M & 1.26T & \sotaa{33.15} & \sotaa{0.9065} & \sotaa{29.29}  & \sotaa{0.7978}   & \sotaa{28.02}  & \sotaa{0.7527} & \sotaa{28.20} & \sotaa{0.8412} & \sotaa{32.63} & \sotaa{0.9306} \\
    \hline
  \end{tabular}
  \end{center}
  \vspace{-3mm}
  \caption{Quantitative comparison (PSNR/SSIM) with state-of-the-art methods on \textbf{classical SR} task. The best and second best results are colored with \sotaa{red} and \sotab{blue}. }
  \label{tab:1}
  \vspace{-4mm}
\end{table*}

\subsection{Experiment Setting}
We use the DF2K dataset, which combines DIV2K~\cite{timofte2017ntire} and Flickr2K~\cite{lim2017enhanced}, as our training set. To ensure fair comparisons, we adopt the same training configuration as utilized in recent SR studies~\cite{chen2023activating, Zhang_2024_CVPR, Tian_2024_CVPR}. Our model is trained for 500K iterations using the AdamW optimizer 
and an initial learning rate of $2 \times 10^{-4}$. The input patch size for training is fixed at $64 \times 64$, and we employ a MultistepLR scheduler, which reduces the learning rate by half at specified iterations $[250000, 400000, 450000, 475000]$. The batch size is set to 32. 
We evaluate our method on five standard benchmark datasets: Set5~\cite{bevilacqua2012low}, Set14~\cite{zeyde2012single}, BSD100~\cite{martin2001database}, Urban100~\cite{huang2015single}, and Manga109~\cite{matsui2017sketch}. In addition, the computational costs of all models in this paper are measured at an output resolution of 1280 $\times$ 640.

\subsection{Comparison with State-of-the-Art Methods}

We compare the performance of our model with various SR baselines on widely used benchmark datasets, including Set5~\cite{bevilacqua2012low}, Set14~\cite{zeyde2012single}, BSD100~\cite{martin2001database}, Urban100~\cite{huang2015single}, and Manga109~\cite{matsui2017sketch}. These SR methods include both classical and state-of-the-art approaches, such as EDSR~\cite{lim2017enhanced}, RCAN~\cite{zhang2018image}, HAN~\cite{niu2020single}, IPT~\cite{Chen_2021_CVPR}, SwinIR~\cite{choi2023n}, CAT~\cite{NEURIPS2022_a37fea8e}, ART~\cite{zhang2023accurate}, HAT~\cite{chen2023activating}, IPG~\cite{Tian_2024_CVPR}, and ATD~\cite{Zhang_2024_CVPR}. We use PSNR and SSIM metrics to assess the performance of SR models at $\times$2, $\times$3, and $\times$4 upscaling factors. All computational costs are calculated at the output resolution of 1280$\times$640, and PFT uses a window size of 32$\times$32.

The results, shown in Tab.~\ref{tab:1}, indicate that the proposed PFT model achieves superior performance over ATD, with significantly reduced computational costs. Notably, PFT reduces computational load by \textbf{17.1\%} compared to ATD, while attaining the highest PSNR in most datasets. For lightweight SR tasks, we further compare our model with other efficient SR methods, including CARN~\cite{ahn2018fast}, IMDN~\cite{hui2019lightweight}, LAPAR~\cite{li2020lapar}, LatticeNet~\cite{luo2020latticenet}, SwinIR~\cite{choi2023n}, SwinIR-NG~\cite{choi2023n}, ELAN~\cite{zhang2022efficient}, and OmniSR~\cite{wang2023omni}. As shown in Tab.~\ref{tab:2}, the proposed PFT-light outperforms the recently introduced ATD-light~\cite{Zhang_2024_CVPR} on nearly all benchmark datasets, while reducing computational complexity by\textbf{ 20.1\%}. On the $\times$2 Urban100 benchmark, PFT-light outperforms ATD-light by \textbf{0.40dB} and IPG-Tiny by \textbf{0.63dB}. For the \( \times 4 \) SR experiments, PFT-light also outperforms IPG Tiny by \textbf{0.42dB} and ATD-light by \textbf{0.23dB} on the Urban100 test set. The superior performance of PFT can be attributed to its use of Progressive Focused Attention, which connects otherwise isolated attention maps across the network. This enables the model to focus on essential features, filter out irrelevant patches, substantially reduce computational costs, and support the use of larger window sizes.

\begin{figure}[htbp]
    \vspace{-4mm}
    \centering
    \includegraphics[width=\linewidth]{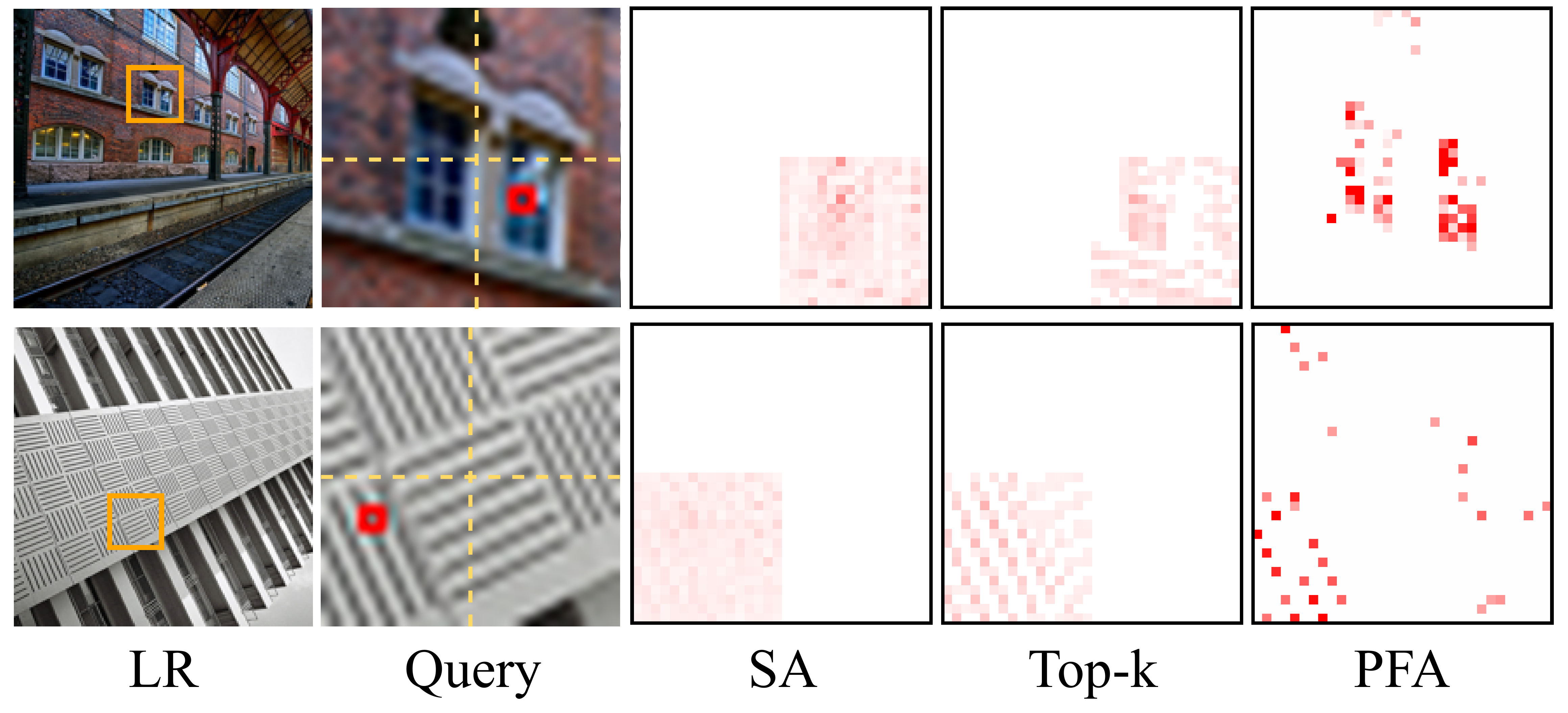}
    \vspace{-8mm}
    \caption{Visual comparison of attention distributions in the 18th layer. SA and top-$k$ Attention distribute attention broadly, failing to focus on the most relevant areas. In contrast, PFA filters out irrelevant tokens and concentrates attention on key regions. By reducing computational costs, it enables the use of a larger 32$\times$32 window for more extensive feature interactions.}
    \vspace{-6mm}
    \label{fig:visual_attention}
\end{figure}

\begin{table*}[htbp]
\captionsetup{font={small}}
\scriptsize
\vspace{-2mm}
  \begin{center}
  \begin{tabular}{|p{2.3cm}|c|c|c|cc|cc|cc|cc|cc|}
    \hline
    \multirow{2}{*}{\textbf{Method}} & \multirow{2}{*}{\textbf{Scale}} & \multirow{2}{*}{\textbf{Params}} & \multirow{2}{*}{\textbf{FLOPs}} & \multicolumn{2}{c|}{\textbf{Set5}} & \multicolumn{2}{c|}{\textbf{Set14}} & \multicolumn{2}{c|}{\textbf{BSD100}} & \multicolumn{2}{c|}{\textbf{Urban100}} & \multicolumn{2}{c|}{\textbf{Manga109}} \\

    & & & & PSNR & SSIM & PSNR & SSIM & PSNR & SSIM & PSNR & SSIM & PSNR & SSIM   \\

    \hline \hline
    CARN~\cite{ahn2018fast}               & $\times$2 & 1,592K & 222.8G   & 37.76 & 0.9590 & 33.52 & 0.9166 & 32.09 & 0.8978 & 31.92 & 0.9256 & 38.36 & 0.9765 \\
    IMDN~\cite{hui2019lightweight}        & $\times$2 & 694K   & 158.8G   & 38.00 & 0.9605 & 33.63 & 0.9177 & 32.19 & 0.8996 & 32.17 & 0.9283 & 38.88 & 0.9774 \\
    LAPAR-A~\cite{li2020lapar}            & $\times$2 & 548K   & 171G   & 38.01 & 0.9605 & 33.62 & 0.9183 & 32.19 & 0.8999 & 32.10 & 0.9283 & 38.67 & 0.9772 \\
    LatticeNet~\cite{luo2020latticenet}   & $\times$2 & 756K   & 169.5G   & 38.15 & 0.9610 & 33.78 & 0.9193 & 32.25 & 0.9005 & 32.43 & 0.9302 & -     & -      \\
    SwinIR-light~\cite{liang2021swinir}   & $\times$2 & 910K   & 244G   & 38.14 & 0.9611 & 33.86 & 0.9206 & 32.31 & 0.9012 & 32.76 & 0.9340 & 39.12 & 0.9783 \\
    ELAN~\cite{zhang2022efficient}        & $\times$2 & 582K   & 203G   & 38.17 & 0.9611 & 33.94 & 0.9207 & 32.30 & 0.9012 & 32.76 & 0.9340 & 39.11 & 0.9782 \\
    SwinIR-NG~\cite{choi2023n}            & $\times$2 & 1181K  & 274.1G   & 38.17 & 0.9612 & 33.94 & 0.9205 & 32.31 & 0.9013 & 32.78 & 0.9340 & 39.20 & 0.9781 \\
    OmniSR~\cite{wang2023omni}            & $\times$2 & 772K   & 194.5G   & 38.22 & 0.9613 & 33.98 & 0.9210 & 32.36 & 0.9020 & 33.05 & 0.9363 & 39.28 & 0.9784 \\    
    IPG-Tiny~\cite{Tian_2024_CVPR}        & $\times$2 & 872K   & 245.2G   & 38.27 & \sotab{0.9616} & \sotaa{34.24} & \sotaa{0.9236} & 32.35 & 0.9018 & 33.04 & 0.9359 & 39.31 & 0.9786 \\    
    ATD-light~\cite{Zhang_2024_CVPR}      & $\times$2 & 753K   & 348.6G   & \sotab{38.28} & \sotab{0.9616} & 34.11 & 0.9217 & \sotab{32.39} & \sotab{0.9023} & \sotab{33.27} & \sotab{0.9376} & \sotab{39.51} & \sotab{0.9789} \\

    \rowcolor{Gray}
    \textbf{PFT-light} (Ours)              & $\times$2 & 776K   & 278.3G   & \sotaa{38.36} & \sotaa{0.9620} & \sotab{34.19} & \sotab{0.9232} & \sotaa{32.43} & \sotaa{0.9030} & \sotaa{33.67} & \sotaa{0.9411} & \sotaa{39.55} & \sotaa{0.9792} \\
    \hline \hline

    CARN~\cite{ahn2018fast}               & $\times$3 & 1,592K & 118.8G & 34.29 & 0.9255 & 30.29 & 0.8407 & 29.06 & 0.8034 & 28.06 & 0.8493 & 33.50 & 0.9440 \\
    IMDN~\cite{hui2019lightweight}        & $\times$3 & 703K   & 71.5G & 34.36 & 0.9270 & 30.32 & 0.8417 & 29.09 & 0.8046 & 28.17 & 0.8519 & 33.61 & 0.9445 \\
    LAPAR-A~\cite{li2020lapar}            & $\times$3 & 544K   & 114G & 34.36 & 0.9267 & 30.34 & 0.8421 & 29.11 & 0.8054 & 28.15 & 0.8523 & 33.51 & 0.9441 \\
    LatticeNet~\cite{luo2020latticenet}   & $\times$3 & 765K   & 76.3G & 34.53 & 0.9281 & 30.39 & 0.8424 & 29.15 & 0.8059 & 28.33 & 0.8538 & -     & -      \\
    SwinIR-light~\cite{liang2021swinir}   & $\times$3 & 918K   & 111G & 34.62 & 0.9289 & 30.54 & 0.8463 & 29.20 & 0.8082 & 28.66 & 0.8624 & 33.98 & 0.9478 \\
    ELAN~\cite{zhang2022efficient}        & $\times$3 & 590K   & 90.1G & 34.61 & 0.9288 & 30.55 & 0.8463 & 29.21 & 0.8081 & 28.69 & 0.8624 & 34.00 & 0.9478 \\
    SwinIR-NG~\cite{choi2023n}            & $\times$3 & 1190K  & 114.1G & 34.64 & 0.9293 & 30.58 & 0.8471 & 29.24 & 0.8090 & 28.75 & 0.8639 & 34.22 & 0.9488 \\
    OmniSR~\cite{wang2023omni}            & $\times$3 & 780K   & 88.4G & \sotab{34.70} & 0.9294 & 30.57 & 0.8469 & 29.28 & 0.8094 & 28.84 & 0.8656 & 34.22 & 0.9487 \\
    IPG-Tiny~\cite{Tian_2024_CVPR}        & $\times$2 & 878K   & 109.0G & 34.64 & 0.9292 & 30.61 & 0.8470 & 29.26 & 0.8097 & 28.93 & 0.8666 & \sotab{34.30} & 0.9493 \\
    ATD-light~\cite{Zhang_2024_CVPR}      & $\times$3 & 760K   & 154.7G & \sotab{34.70} & \sotab{0.9300} & \sotab{30.68} & \sotab{0.8485} & \sotab{29.32} & \sotab{0.8109} & \sotab{29.16} & \sotab{0.8710} & \sotaa{34.60} & \sotab{0.9505} \\
    
    \rowcolor{Gray}
    \textbf{PFT-light} (ours)              & $\times$3 & 783K   & 123.5G & \sotaa{34.81} & \sotaa{0.9305} & \sotaa{30.75} & \sotaa{0.8493} & \sotaa{29.33} & \sotaa{0.8116} & \sotaa{29.43} & \sotaa{0.8759} & \sotaa{34.60} & \sotaa{0.9510} \\

    \hline \hline
    CARN~\cite{ahn2018fast}               & $\times$4 & 1,592K & 90.9G & 32.13 & 0.8937 & 28.60 & 0.7806 & 27.58 & 0.7349 & 26.07 & 0.7837 & 30.47 & 0.9084 \\
    IMDN~\cite{hui2019lightweight}        & $\times$4 & 715K   & 40.9G & 32.21 & 0.8948 & 28.58 & 0.7811 & 27.56 & 0.7353 & 26.04 & 0.7838 & 30.45 & 0.9075 \\
    LAPAR-A~\cite{li2020lapar}            & $\times$4 & 659K   & 94G & 32.15 & 0.8944 & 28.61 & 0.7818 & 27.61 & 0.7366 & 26.14 & 0.7871 & 30.42 & 0.9074 \\
    LatticeNet~\cite{luo2020latticenet}   & $\times$4 & 777K   & 43.6G & 32.30 & 0.8962 & 28.68 & 0.7830 & 27.62 & 0.7367 & 26.25 & 0.7873 & -     & -      \\
    SwinIR-light~\cite{liang2021swinir}   & $\times$4 & 930K   & 63.6G & 32.44 & 0.8976 & 28.77 & 0.7858 & 27.69 & 0.7406 & 26.47 & 0.7980 & 30.92 & 0.9151 \\
    ELAN~\cite{zhang2022efficient}        & $\times$4 & 582K   & 54.1G & 32.43 & 0.8975 & 28.78 & 0.7858 & 27.69 & 0.7406 & 26.54 & 0.7982 & 30.92 & 0.9150 \\
    SwinIR-NG~\cite{choi2023n}            & $\times$4 & 1201K  & 63.0G & 32.44 & 0.8980 & 28.83 & 0.7870 & 27.73 & 0.7418 & 26.61 & 0.8010 & 31.09 &0.9161 \\
    OmniSR~\cite{wang2023omni}            & $\times$4 & 792K   & 50.9G & 32.49 & 0.8988 & 28.78 & 0.7859 & 27.71 & 0.7415 & 26.65 & 0.8018 & 31.02 & 0.9151 \\
    IPG-Tiny~\cite{Tian_2024_CVPR}       & $\times$4 & 887K   & 61.3G & 32.51 & 0.8987 & 28.85 & 0.7873 & 27.73 & 0.7418 & 26.78 & 0.8050 & 31.22 & 0.9176 \\
    ATD-light~\cite{Zhang_2024_CVPR}       & $\times$4 & 769K   & 87.1G & \sotab{32.62} & \sotab{0.8997} & \sotab{28.87} & \sotab{0.7884} & \sotab{27.77} & \sotab{0.7439} & \sotab{26.97} & \sotab{0.8107} & \sotab{31.47} & \sotab{0.9198} \\
    
    \rowcolor{Gray}
    \textbf{PFT-light} (Ours)              & $\times$4 & 792K   & 69.6G   & \sotaa{32.63} & \sotaa{0.9005} & \sotaa{28.92} & \sotaa{0.7891} & \sotaa{27.79} & \sotaa{0.7445} & \sotaa{27.20} & \sotaa{0.8171} & \sotaa{31.51} & \sotaa{0.9204} \\
    \hline
  \end{tabular}
  \end{center}
  \vspace{-5mm}
  \caption{Quantitative comparison (PSNR/SSIM) with state-of-the-art methods on \textbf{lightweight SR} task. The best and second best results are colored with \textcolor{red}{red} and \textcolor{blue}{blue}.}
\vspace{-4mm}
\label{tab:2}
\end{table*}


\noindent \textbf{Visual comparison of attention distribution}. The attention distribution visualizations are shown in Fig.~\ref{fig:visual_attention}.
The attention distributions of the Vanilla Self-Attention and top-$k$ Attention methods are relatively scattered, failing to allocate attention to the regions most relevant to the current query. In contrast, our PFA method filters out irrelevant tokens and focuses on most important positions. It also lowers computation costs, enabling larger 32$\times$32 windows for broader feature interactions.

\begin{figure*}[htbp]
    \centering
    \includegraphics[width=\linewidth]{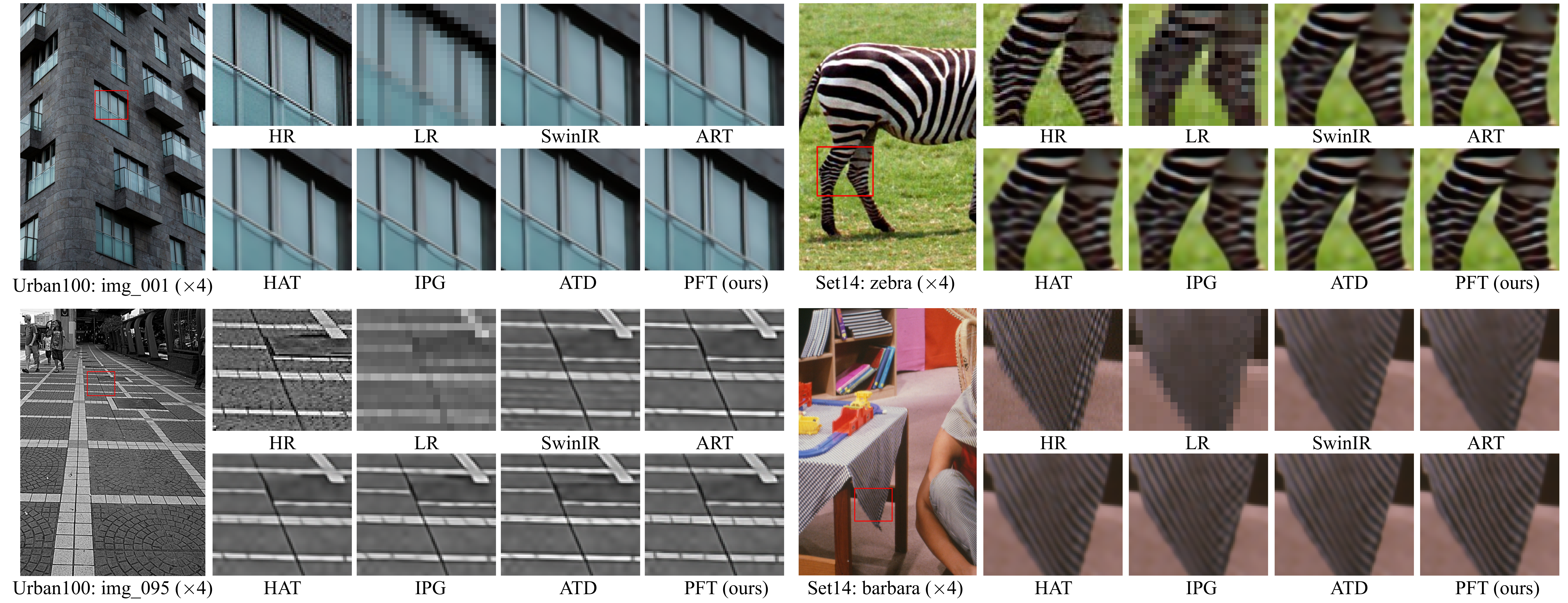}
    \vspace{-4mm}
    \caption{Visual comparison of SR reconstruction results.}
    \label{fig:visual}
\end{figure*}

\noindent \textbf{Visual comparison of SR reconstruction results}.~To evaluate the quality of different super-resolution methods, we show visual examples in Fig.~\ref{fig:visual}. These comparisons demonstrate the strength of our PFT approach in restoring sharp edges and fine textures from low-resolution images. For example, in img\_001 and img\_095, most methods struggle to reconstruct textures accurately, leading to noticeable distortions. Similarly, in the zebra and barbara images, many methods fail to recover details, resulting in blurry or incorrect textures. In contrast, our PFT effectively captures fine textures by focusing on the most relevant features. This ability to selectively attend helps restore clean edges and reduce artifacts, producing more accurate reconstructions.

\subsection{Ablation Study}

We conduct ablation studies on the proposed PFT-light model, with all models trained for 250k iterations on the DIV2K dataset at $\times 4$ scale. Except for the experiments on window size, all other experiments use a window size of $16 \times 16$. In top-$k$ Attention, the value of $k$ is set to half the number of window elements, which is 128.

\noindent \textbf{Effectiveness of Progressive Focused Attention}. We conducted experiments to demonstrate the effectiveness of the proposed Progressive Attention and Progressive Focused Attention, with results presented in Tab.~\ref{tab:effectiveness}. These experiments compare the quantitative performance of $\times$4 super-resolution on the Set5, Urban100, and Manga109 datasets. Progressive Attention applies a straightforward Hadamard product connection of the attention maps. As shown, even the Progressive Attention provides notable performance improvements over Self-Attention and top-$k$ Attention. Progressive Focused Attention achieved the best results with a \textbf{27.69\%} reduction in computation, 
and notably increasing PSNR by \textbf{0.36dB} on the Urban100 dataset. The effectiveness of PFA lies in its ability to link adjacent attention maps across the network via the Hadamard product, allowing the model to reduce redundant computations and focus on the most relevant tokens.

\begin{table}[htbp]
\captionsetup{}
\vspace{-2mm}
  \centering
  \begin{center}
  \resizebox{\linewidth}{!}{
  \begin{tabular}{l|ccccc}
    \toprule[1.1pt]
    Method                          & FLOPs & Set5 & Urban100 & Manga109 \\

    \midrule
    Vanilla Self-Attention          & 70.4G & 32.28 & 26.26  & 30.62  \\
    Top-$k$ Attention                 & 70.4G & 32.30 & 26.29  & 30.67  \\
    Progressive Attention           & 70.4G & 32.35 & 26.41  & 30.78  \\
    Progressive Focused Attention   & 50.9G & 32.41 & 26.62  & 30.85  \\  
    \bottomrule[1.1pt]
  \end{tabular}}
  \end{center}
  \vspace{-5mm}
  \caption{Ablation study on the proposed focused attention.}
  \vspace{-2mm}
  \label{tab:effectiveness}
\end{table}

\noindent \textbf{Effects of focus ratio}. As discussed in Sec.~\ref{PFA}, the parameter $\alpha$ controls the focus degree of Progressive Focused Attention as network depth increases. A smaller $\alpha$ retains fewer attention positions, accelerating the focusing process. However, $\alpha$ is generally kept above 0.1, as a rapid reduction in attention retention may cause the network to make premature choices. As shown in Tab.~\ref{tab:ratio}, model performance is optimal when $\alpha$ is around 0.5. Experimental results indicates that both excessively large and small values of $\alpha$ can hinder effective attention focusing. 
A very small $\alpha$ may cause the network to overlook important tokens, leading to incomplete representations. On the other hand, a large $\alpha$ keeps too many irrelevant tokens, weakening the focus and limiting the network's ability to reduce noise.

\begin{table}[htbp]
\captionsetup{}
\vspace{-1mm}
  \centering
  \begin{center}
  \resizebox{0.97\linewidth}{!}{
  \begin{tabular}{l|ccccc}
    \toprule[1.pt]
    Focus ratio & 0.1 & 0.3 & 0.5 & 0.7 & 0.9 \\
    \midrule
    PSNR   & 26.48 & 26.58 & 26.62 & 26.61 & 26.56 \\
    \bottomrule[1.pt]
  \end{tabular}}
  \end{center}
  \vspace{-4mm}
  \caption{Ablation study on focus ratio for \( \times 4 \) SR, evaluated on Urban100.}
  \vspace{-1mm}
  \label{tab:ratio}
\end{table}

\noindent \textbf{Effects of different window sizes}. The window size determines the model's scope of correlation extraction and significant impact on reconstruction performance. Previous studies have shown that larger windows yield better reconstruction results for image super-resolution tasks. However, due to the quadratic complexity of self-attention, previous methods find it challenging to use large windows effectively. Our PFA dynamically filters out patches irrelevant to the current query patch. This allows us to select the interaction scope before computing similarity. As a result, the overall computational load is greatly reduced. This enables us to use large windows while maintaining a low computational cost. As shown in Tab.~\ref{tab:win_size}, reconstruction performance improves as the window size increases. We ultimately select a window size of $32 \times 32$ to balance performance, model parameters, and computational cost.

\begin{table}[htbp]
\captionsetup{}
\scriptsize
\vspace{-2mm}
  \centering
  \begin{center}
  \resizebox{1\linewidth}{!}{
  \begin{tabular}{l|cccc}
    \toprule[0.7pt]
    Window size & Set5 & Urban100 & Manga109  \\
    \midrule
    8 $\times$ 8      & 32.33 & 26.40 & 30.71 &  \\
    16 $\times$ 16    & 32.41 & 26.62 & 30.85 \\
    32 $\times$ 32    & 32.49 & 26.81 & 30.93 \\  
    \bottomrule[0.7pt]
  \end{tabular}}
  \end{center}
  \vspace{-5mm}
  \caption{Ablation study on window sizes for $\times$4 SR, evaluated on the Urban100 dataset, with PSNR as the evaluation metric.}
  \vspace{-6mm}
  \label{tab:win_size}
\end{table}

\section{Conclusion}
\label{sec:conclusion}

%
We propose the PFT, which introduces a novel approach to linking attention maps across layers through the key innovation of the PFA mechanism, progressively refining the focus on relevant features.
By inheriting and modifying attention maps from previous layers, the PFA mechanism reduces unnecessary similarity calculations, improving attention efficiency.
This enables the model to prioritize the most relevant tokens while effectively suppressing the influence of irrelevant ones. 
%
%
%
%
Experiments show that PFT and PFT-light achieves leading performance in single-image super-resolution with lower computation. 
%
%
As a key component of the Progressive Focused Transformer, PFA has proven to be both effective and efficient for super-resolution tasks. We look forward to exploring its potential in other high-level vision tasks, as well as in natural language processing.

\section*{Acknowledgement}
This work was supported by the National Natural Science Foundation of China (No.~62476051) and the Sichuan Natural Science Foundation (No.~2024NSFTD0041).

{
    \small
    \bibliographystyle{ieeenat_fullname}
    \bibliography{main}
}

\clearpage
\setcounter{page}{1}
\maketitlesupplementary

In this supplementary material, we provide additional details on model training, inference time efficiency comparisons, and more comprehensive visual results. Specifically, in Section A, we present the training details for the PFT and PFT-light models. Subsequently, in Section B, we compare the inference time efficiency of different models. Finally, in Section C, we provide more detailed visualizations of the model's results.

\section*{A. Training Details}
\label{sec:rationale}

For training the PFT model, we use the DF2K dataset, which combines DIV2K~\cite{timofte2017ntire} and Flickr2K~\cite{lim2017enhanced}, as our training set. To ensure fair comparisons, we adopt the same training configurations as those employed in recent super-resolution (SR) studies~\cite{chen2023activating, Zhang_2024_CVPR, Tian_2024_CVPR}. Our model is optimized using the AdamW optimizer with parameters set to \((\beta_1 = 0.9, \beta_2 = 0.99)\), a weight decay coefficient \(\lambda = 0.0001\), and an initial learning rate of \(2 \times 10^{-4}\). The \(\times 2\) model is trained for 500K iterations. During training, the input patch size is fixed at \(64 \times 64\), and a MultistepLR scheduler is applied to halve the learning rate at predefined iterations \([250000, 400000, 450000, 475000]\). The batch size is set to 32 for all training processes. To enhance robustness, the training data is augmented with random horizontal and vertical flips as well as random rotations of \(90^\circ\). For the \(\times 3\) and \(\times 4\) models, we apply fine-tuning based on the pre-trained \(\times 2\) model to save time, training these models for only 250K iterations. The initial learning rate is set to \(2 \times 10^{-4}\), and a MultistepLR scheduler is used to halve the learning rate at predefined iterations \([100000, 150000, 200000, 225000, 240000]\). We evaluate our method on five standard benchmark datasets: Set5~\cite{bevilacqua2012low}, Set14~\cite{zeyde2012single}, BSD100~\cite{martin2001database}, Urban100~\cite{huang2015single}, and Manga109~\cite{matsui2017sketch}. Additionally, the computational cost of all models presented in this paper is measured at an output resolution of \(1280 \times 640\). 
For training the PFT-light model, only the DIV2K dataset is used, excluding Flickr2K. The initial learning rate for training \(\times 2\) SR is set to \(5 \times 10^{-4}\). All other training strategies remain consistent with those used for the PFT model.

\section*{B. Comparison of inference time}
\label{sec:inference_time}

We compare the inference time of our PFT model with several state-of-the-art SR methods, including HAT~\cite{chen2023activating}, IPG~\cite{Tian_2024_CVPR}, and ATD~\cite{Zhang_2024_CVPR}. In this experiment, the inference time for all models is measured on a single NVIDIA GeForce RTX 4090 GPU at an output resolution of \(512 \times 512\). As shown in Tab.~\ref{tab:time}, the inference time of our PFT model is comparable to existing methods. At the \(\times 2\) and \(\times 3\) scales, our model takes more time than HAT and ATD but is significantly faster than IPG. At the \(\times 4\) scale, PFT outperforms both ATD and IPG in terms of inference speed. This improvement can be attributed to the efficient SMM CUDA kernels we developed to accelerate sparse matrix multiplication. Notably, despite the minor differences in inference time, our PFT model achieves lower computational complexity and delivers the best reconstruction performance.

\begin{table}[htbp]
    \vspace{-2mm}
    \centering
    \resizebox{\linewidth}{!}{
        \begin{tabular}{l|l|ccccc}
            \toprule[1.1pt]
            Scale & Method & Params & FLOPs & Inference time	\\
            \midrule
            \multirow{4}{*}{$\times$ 2} 
            & HAT~\cite{chen2023activating}   & 20.6M & 5.81T & 1078ms  \\
            & ATD~\cite{Zhang_2024_CVPR}      & 20.1M & 6.07T & 1394ms  \\
            & IPG~\cite{Tian_2024_CVPR}       & 18.1M & 5.35T & 2320ms  \\
            
            \rowcolor{Gray}
            & PFT (Ours)                      & 19.6M & 5.03T & 1594ms   \\
            
            \hline
            \multirow{4}{*}{$\times$ 3} 
            & HAT~\cite{chen2023activating}   & 20.8M & 2.58T & 799ms \\
            & ATD~\cite{Zhang_2024_CVPR}      & 20.3M & 2.69T & 1038ms  \\
            & IPG~\cite{Tian_2024_CVPR}       & 18.3M & 2.39T & 1651ms  \\
            
            \rowcolor{Gray}
            & PFT (Ours)                      & 19.8M & 2.23T & 1158ms   \\

            \hline
            \multirow{4}{*}{$\times$ 4} 
            & HAT~\cite{chen2023activating}   & 20.8M & 1.45T & 725ms  \\
            & ATD~\cite{Zhang_2024_CVPR}      & 20.3M & 1.52T & 867ms  \\
            & IPG~\cite{Tian_2024_CVPR}       & 17.0M & 1.30T & 1060ms  \\
            
            \rowcolor{Gray}
            & PFT (Ours)                      & 19.8M & 1.26T & 852ms   \\

            \bottomrule[1.1pt]
    \end{tabular}}
    \vspace{-2mm}
    \caption{Inference efficiency comparison of models.}
    \vspace{-6mm}
    \label{tab:time}
\end{table}

\section*{C. More Visual Examples.}
\label{sec:m_visual}

\subsection*{C.1. Visual of attention distributions.}
\label{sec:visual_attention_dis}

\begin{figure*}[htbp]
    \vspace{-6mm}
    \centering
    \includegraphics[width=\linewidth]{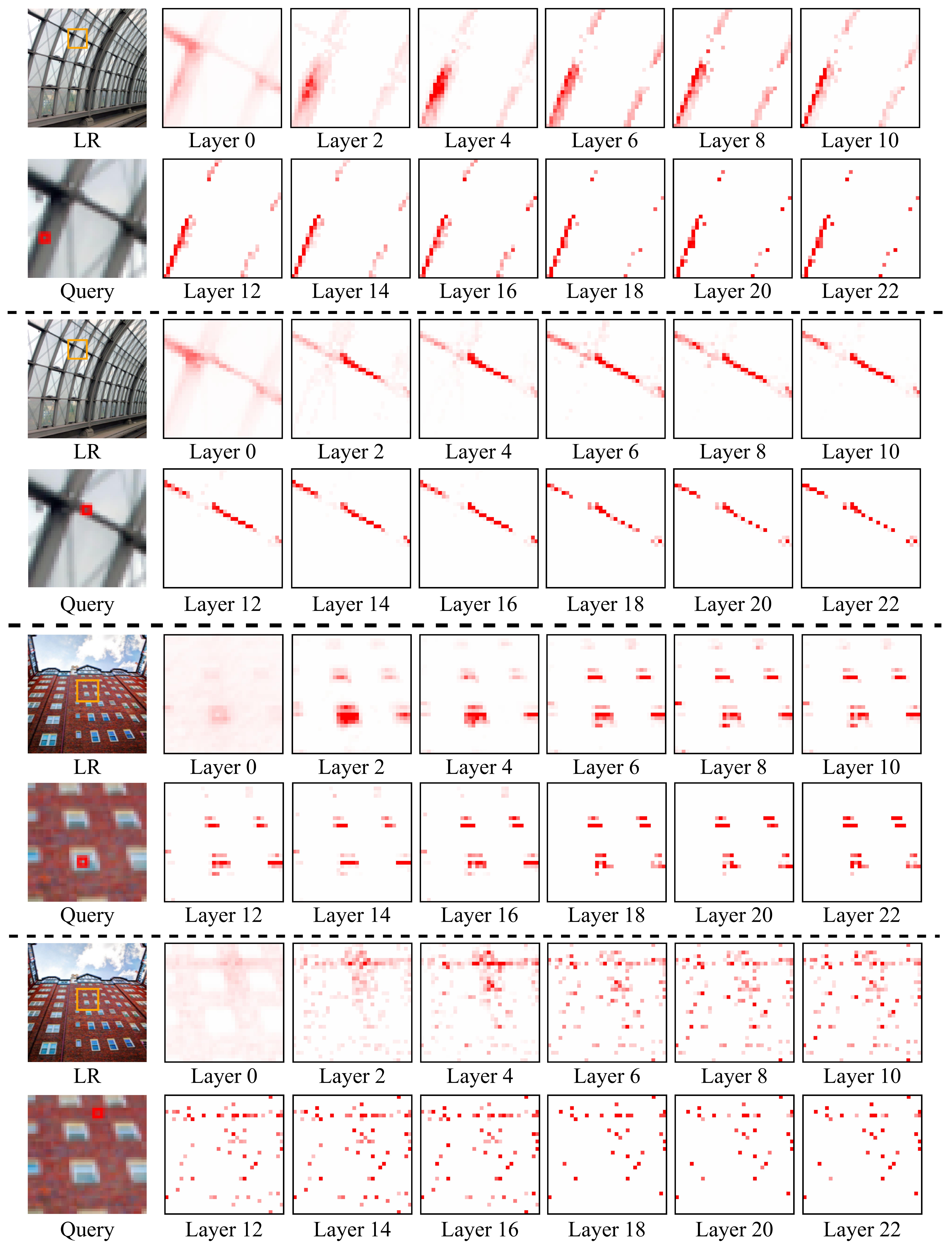}
    \vspace{-5mm}
    \caption{The visualization of attention distributions across different layers of the PFT-light model demonstrates the progressive filtering capability of the PFA module.}
    \vspace{-1mm}
    \label{fig:visual_supp_attention_map}
    \vspace{-3mm}
\end{figure*}

The visualization of attention distributions across different layers of the PFT-light model is shown in Fig.~\ref{fig:visual_supp_attention_map}. As the network deepens, the PFA module progressively filters out tokens irrelevant to the current query and concentrates attention on the most critical regions. This mechanism not only reduces the influence of irrelevant features on reconstruction performance but also lowers computational costs, enabling the model to perform feature interactions over a larger spatial scope.

\subsection*{C.2. Visual Comparisons of PFT-light.}
\label{sec:visual_PFT_light}
\begin{figure*}[htbp]
    \vspace{-6mm}
    \centering
    \includegraphics[width=\linewidth]{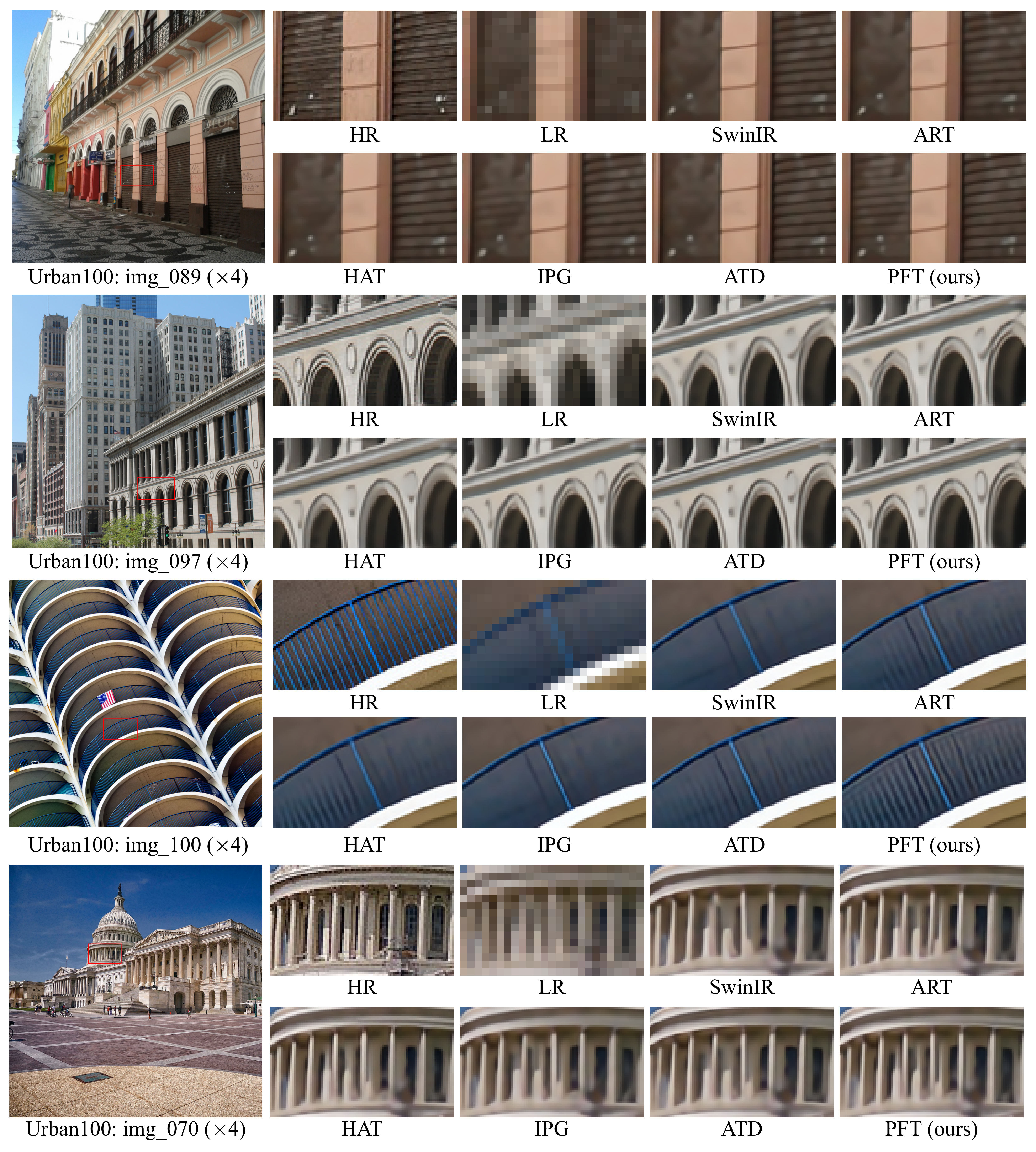}
    \vspace{-5mm}
    \caption{Visual comparison of classical SR reconstruction results.}
    \vspace{-1mm}
    \label{fig:visual_supp1}
    \vspace{-3mm}
\end{figure*}

\begin{figure*}[htbp]
    \vspace{-6mm}
    \centering
    \includegraphics[width=\linewidth]{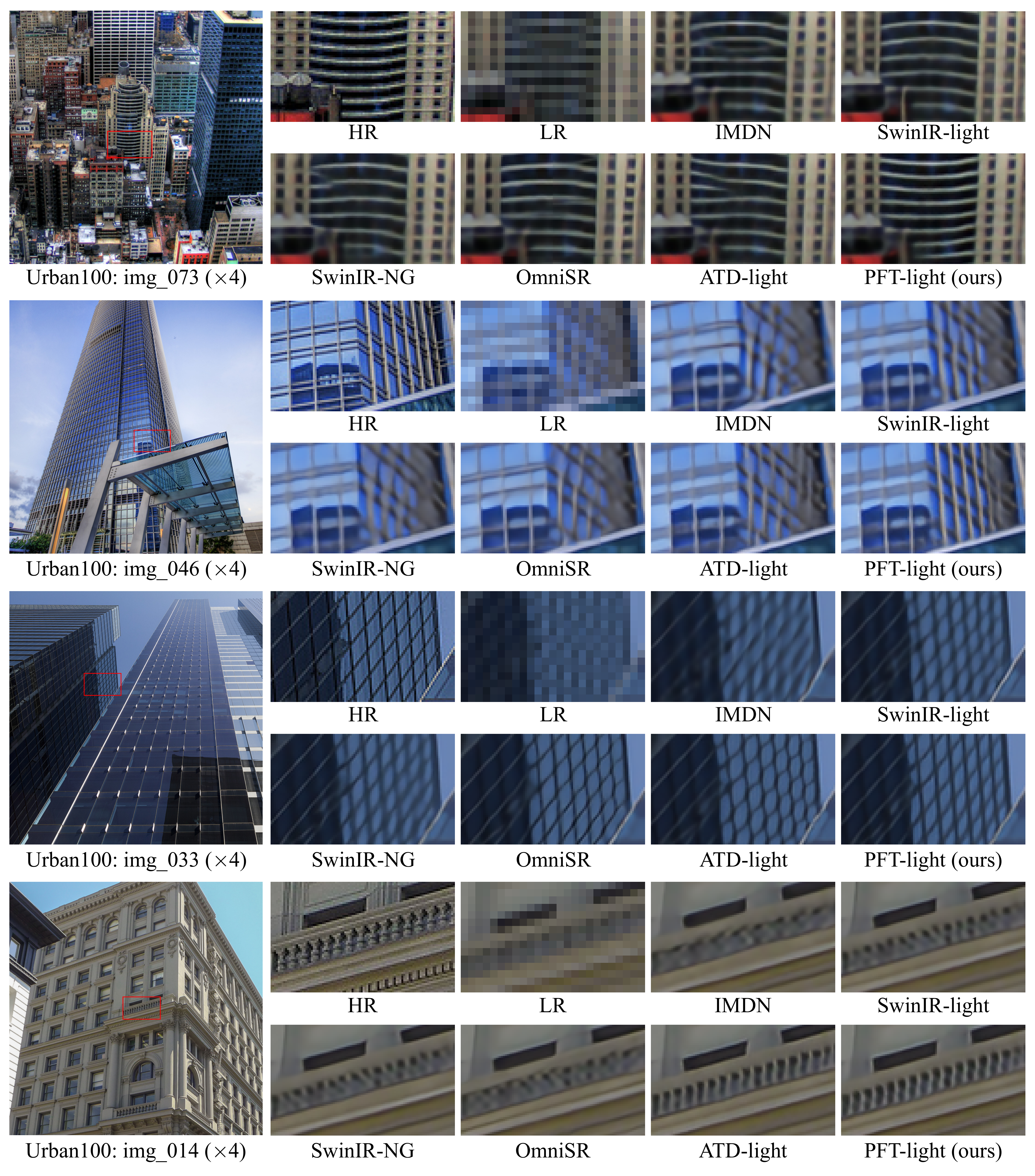}
    \vspace{-5mm}
    \caption{Visual comparison of lightweight SR reconstruction results.}
    \vspace{-1mm}
    \label{fig:visual_supp2}
    \vspace{-3mm}
\end{figure*}

\begin{figure*}[htbp]
    \vspace{-6mm}
    \centering
    \includegraphics[width=\linewidth]{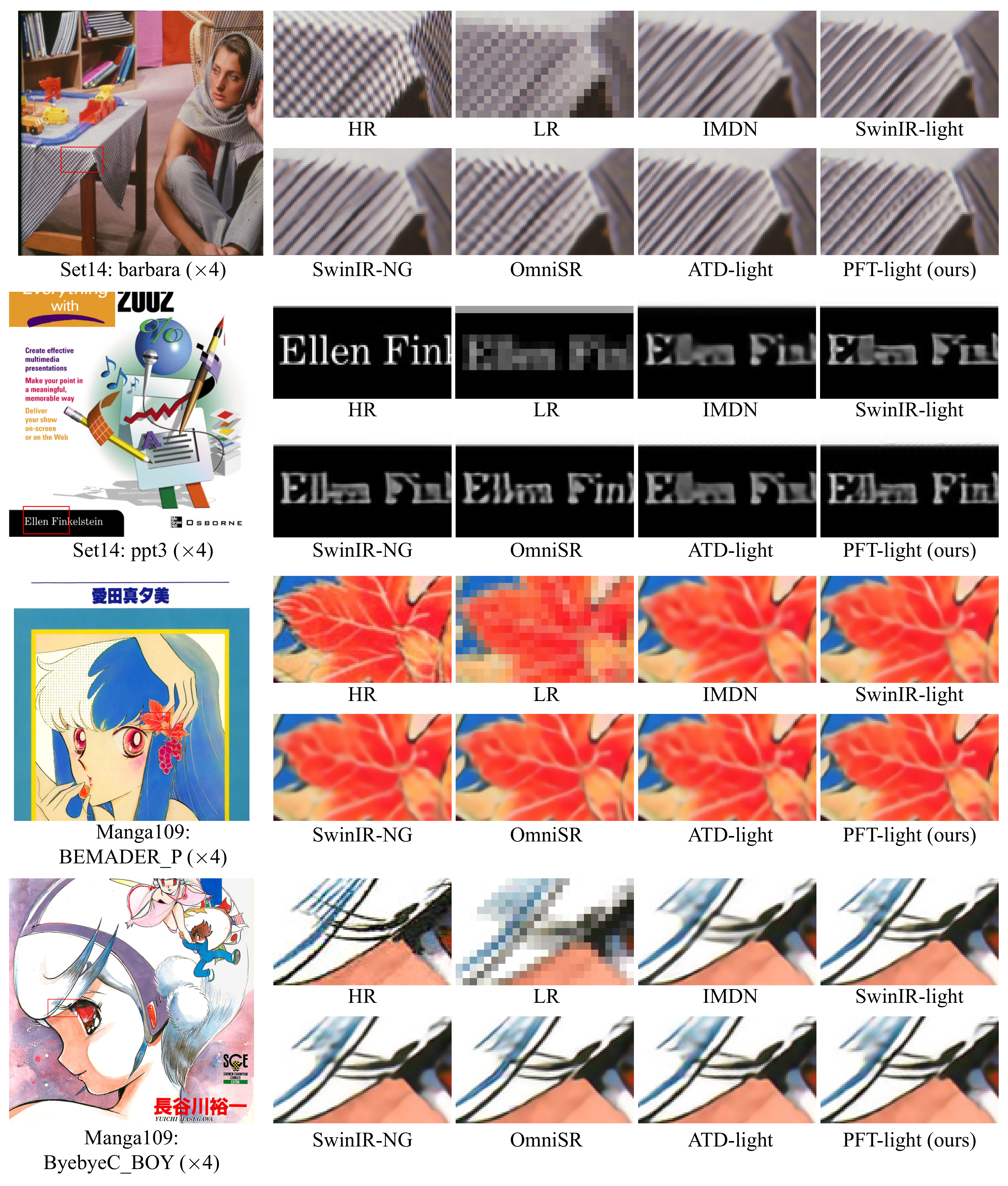}
    \vspace{-5mm}
    \caption{Visual comparison of lightweight SR reconstruction results.}
    \vspace{-1mm}
    \label{fig:visual_supp3}
    \vspace{-3mm}
\end{figure*}

To qualitatively evaluate the reconstruction performance of our PFT and PFT-light models in comparison with other methods, we provide visual examples in Fig.~\ref{fig:visual_supp1}, Fig.~\ref{fig:visual_supp2}, and Fig.~\ref{fig:visual_supp3}. These comparisons emphasize the strengths of our approach in restoring sharp edges and fine textures from severely degraded low-resolution inputs. The PFT-light model, in particular, excels at capturing edge details. Its selective focus on critical regions allows it to produce cleaner edges and achieve more accurate and visually reasonable reconstructions.

\end{document}